\documentclass[conference]{IEEEtran}
\IEEEoverridecommandlockouts

\usepackage{cite}
\usepackage{amsmath,amssymb,amsfonts}
\usepackage{algorithmic}
\usepackage{graphicx}
\usepackage{textcomp}
\usepackage{booktabs}
\usepackage{graphicx}
\usepackage{tabularx}
\usepackage{makecell}
\usepackage{tikz}
\usetikzlibrary{arrows.meta}
\usepackage{xcolor}
\def\BibTeX{{\rm B\kern-.05em{\sc i\kern-.025em b}\kern-.08em
    T\kern-.1667em\lower.7ex\hbox{E}\kern-.125emX}}

\begin{document}

\title{Spatially-Aware Class-Agnostic Object Counting\\
\thanks{$^{*}$Corresponding email: robert\_wijaya@mymail.sutd.edu.sg.}}

\author{%
\IEEEauthorblockN{%
Robert Wijaya\textsuperscript{1*}, Md. Tanvir Hossain\textsuperscript{2}, Amanda Kau\textsuperscript{3}, Ngai-Man Cheung\textsuperscript{1} %
}%
\IEEEauthorblockA{\small%
\textsuperscript{1}Temasek Laboratories, Singapore University of Technology and Design, Singapore\\
\textsuperscript{2}The University of Canterbury, Christchurch, New Zealand\\
\textsuperscript{3}The Australian National University, Canberra, Australia%
\\
}%
}
\maketitle

\begin{abstract}
Generalised object counting aims to estimate the number of instances of an arbitrary object category from a single image, but many recent methods can struggle on structurally complex objects due to limited spatial modelling. We present \textit{UpCount}, a class-agnostic counter designed to better preserve spatial structure. UpCount strengthens the visual representation by extracting multi-layer features from a ViT-B/16 encoder and reassembling them into a refined multi-scale pyramid that is spatially refined using Dense Prediction Transformers and FeatUp, yielding features with improved structural and spatial sensitivity; a proposal--verification counting head then identifies repeated patterns and produces a density map for the final count. On FSC-147, UpCount achieves 12.39 MAE and 100.89 RMSE on the test set, and it transfers effectively to vehicle counting on CARPK (6.27 MAE, 8.79 RMSE). Code: \texttt{https://github.com/r28112072-rgb/upcount}.
\end{abstract}


\section{Introduction}
Object counting in computer vision aims to accurately quantify any particular object in an image, 
and is applicable to logistics, surveillance, urban planning and more \cite{xu2023zero, huang2024pointsegmentcountgeneralized, hobley2024abceasy123blind, hobley2022learningcountanythingreferenceless, 10656642, Ranjan_2022_ACCV}.
Specialised counters such as ones for crowd-counting  \cite{Zhang_2016_CVPR} and vehicles \cite{mundhenk2016large} exist.
These specialised counters can effectively count large quantities of objects but are limited to specific object classes, 
a limitation that the human counting ability does not possess.
Other general counters can count arbitrary object classes, but require additional text prompts or exemplars.
Exemplars are particularly inefficient as they require humans to manually produce annotations like bounding boxes to isolate the desired object to count.

To this end, Liu et al. \cite{countr} developed Counting TRansformer (CounTR), a transformer-based model that can count any arbitrary class of objects in an effort to mimic our human counting ability.
The authors were motivated by findings from Lu et al. in \cite{lu2019class}, where the object counting problem was posed as a matching problem for image patches, and self-similarity was an found to be an important element for counting.
Attention mechanisms were built into CounTR's architecture as a result, 
which seemed to give CounTR an edge over several state-of-the-art (SOTA) models.

However, CounTR, like many other models, failed where object structural information was key.
For instance, the counts of sunglasses would be double the true count as each lens would be treated as a separate object.
These cases necessitated test-time augmentation, which involved performing transformations to the query image to obtain several augmented images, performing counting on each image, then returning the average count as the final one.
This introduced additional steps into the pipeline and increased computational demands. We hypothesize that this limitation may stem from the feature interaction module in CounTR \cite{countr}, which did not sufficiently capture structural information. 

We address this limitation by strengthening the visual representation at the very first stage of the pipeline. Specifically, we replace the original encoder with an MAE-pretrained ViT-B/16 and extract features from multiple intermediate Transformer blocks, which are then reassembled into a multi-scale pyramid and spatially refined (Dense Prediction Transformers (DPT) \cite{ranftl2021vision} + FeatUp \cite{fu2024featup}). We call this model \textit{UpCount}. To convert these spatially refined features into final predictions, we adopt a proposal--verification counting head that detects repeated visual patterns and outputs a density map and count. This design preserves fine structural cues while retaining high-level semantics, improving zero-shot counting of structurally complex objects, while also eliminating the exemplar boxes or text prompts at inference time.

In the reference-free setting (no exemplar boxes or text prompts at inference), UpCount achieves strong performance on FSC-147 with a validation MAE/RMSE of 13.62/60.07 and a test MAE/RMSE of 12.39/100.89. On CARPK, UPCount reaches 6.27 MAE and 8.79 RMSE, substantially improving over exemplar-guided methods.

Our main contributions are as follows:
\begin{itemize}
    \item We propose \textit{UpCount}, a reference-free, class-agnostic counting model with strong spatial generalization that requires no exemplar boxes, or text prompts at inference.
    \item We strengthen the visual representation by extracting multi-layer ViT features and reassembling them into a refined multi-scale pyramid (DPT \cite{ranftl2021vision} + FeatUp \cite{fu2024featup}) for improved structural and spatial sensitivity.
    \item We validate UpCount on FSC-147 \cite{ranjan2021learning} and CARPK, showing improved performance over representative reference-free baselines and strong transfer to vehicle counting.
\end{itemize}

\begin{figure*}[htbp]
\centering
\includegraphics[width=\textwidth,height=0.3\textheight,keepaspectratio]{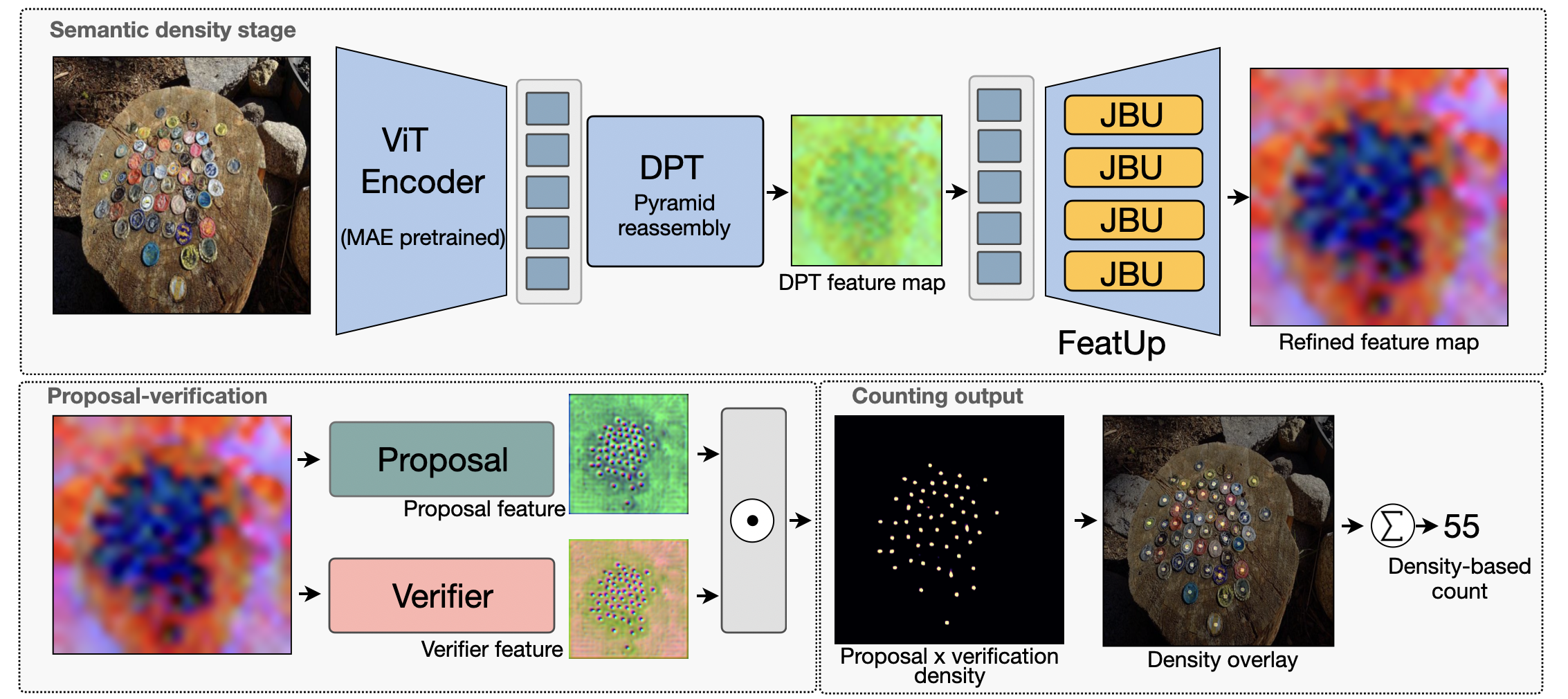}
\caption{Overview of the proposed UpCount architecture.}
\label{arch}
\end{figure*}

\section{Related Work}
\subsection{Class-specific Object Counting}
Object counting is commonly grouped into class-specific and class-agnostic settings. Class-specific methods target predefined categories (e.g., people \cite{Zhang_2016_CVPR} or cars \cite{mundhenk2016large}) and are typically either detection-based (counting bounding boxes) or regression-based (summing a predicted density map) \cite{xu2023zero}. Their reliance on fixed classes limits extension to arbitrary or multiple object types within an image.

\subsection{Class-agnostic Object Counting}
Early class-agnostic counters such as GMN \cite{lu2019class} and FamNet \cite{ranjan2021learning} rely on annotated exemplars and substantial training data (e.g., GMN leverages large-scale ILSVRC detection videos), which can be impractical to collect. Exemplar-free approaches reduce this burden but often introduce other constraints, e.g., RepRPN \cite{ranjan2022exemplar} primarily counts dominant classes, while text-conditioned methods generate or select class-relevant regions from prompts (VAE-based \cite{xu2023zero}, CLIP-Count \cite{jiang2023clip}, CounTX \cite{countx} using CLIP \cite{CLIP}), still requiring user-specified text and limiting fully automatic counting.

\section{Method}
\subsection{Architecture}

We propose a spatially-aware, class-agnostic object counting model that predicts a full-resolution density map and the corresponding count without category labels, text prompts, or exemplar boxes. It detects repeated patterns directly from the input image built from four main components:
\subsubsection{\textbf{MAE-Pretrained ViT Encoder}} We employ a vanilla ViT-B/16 encoder initialized through masked autoencoding. Given an input image
$\mathcal{X}_i \in \mathbb{R}^{3 \times 384 \times 384}$, the image is
partitioned into non-overlapping $16 \times 16$ patches, producing a
$24 \times 24$ grid of $M=576$ visual tokens. 

The encoder consists of 12 Transformer blocks with 12 attention heads. Instead of using only the output of the final block, we extract layer-normalized representation from four intermediate depths:
\begin{equation}
    \mathbf{Z}^{(l)}
    =
    \Phi_{\mathrm{ViT\text{-}Enc}}^{(l)}
    \left(\mathcal{X}_i\right)
    \in \mathbb{R}^{M \times D},
    \qquad
    l \in \{2,5,8,11\}.
\end{equation} where $l$ denotes the zero-based Transformer-block index. Thus, the selected indices correspond to the 3rd, 6th, 9th, and 12th Transformer blocks, respectively. The complete multi-level encoder output is written as
\begin{equation}
    \mathcal{F}_{\mathrm{ViT}}
    =
    \left\{
    \mathbf{Z}^{(2)},
    \mathbf{Z}^{(5)},
    \mathbf{Z}^{(8)},
    \mathbf{Z}^{(11)}
    \right\},
    \qquad
    \mathbf{Z}^{(l)}
    \in
    \mathbb{R}^{576 \times 768}.
\end{equation} Each token sequence is subsequently reshaped into a spatial feature map
\begin{equation}
\mathbf{F}^{(l)}
=
\mathcal{R}\left(\mathbf{Z}^{(l)}\right)
\in
\mathbb{R}^{D \times 24 \times 24},
\end{equation} where $\mathcal{R}(\cdot)$ denotes the token-to-grid reshaping operation. The resulting spatial feature maps retain representations at different semantic depths while sharing the same spatial resolution.

\subsubsection{\textbf{DPT Pyramid Reassembly}} 
All intermediate ViT features have the same $24 \times 24$ spatial resolution. To construct a hierarchical representation, we employ a DPT-style pyramid reassembly module \cite{ranftl2021vision}. The projected features from blocks $\{2,5,8,11\}$ are reassembled at resolutions of $96 \times 96$, $48 \times 48$, $24 \times 24$, and $12 \times 12$, respectively.The lower-resolution features are progressively upsampled and combined with the higher-resolution features. The complete operation is expressed as
\begin{equation}
    \mathbf{F}_{\mathrm{DPT}}
    =
    \Phi_{\mathrm{DPT}}
    \left(
        \mathbf{F}^{(2)},
        \mathbf{F}^{(5)},
        \mathbf{F}^{(8)},
        \mathbf{F}^{(11)}
    \right).
\end{equation} The resulting feature map is
\begin{equation}
    \mathbf{F}_{\mathrm{DPT}}
    \in
    \mathbb{R}^{C_o \times 96 \times 96},
\end{equation}
where $C_o$ denotes the channel dimension of the reassembled DPT feature map and is set to 64 in our implementation. In parallel, the projected final block feature is retained at $24 \times 24$ resolution as the semantic input to the subsequent FeatUp refinement module.

\begin{table*}[t]
\caption{Performance comparison on FSC-147}
\label{tab1}
\centering
\begin{tabular}{|c|c|c|c|c|c|c|c|}
\hline
\textbf{Method} &  \textbf{Year} & \textbf{Inference Guidance} & \textbf{\# Shots} & \textbf{Val MAE} & \textbf{Val RMSE} & \textbf{Test MAE} & \textbf{Test RMSE} \\
\hline
T2ICount & 2025 &Text & 0 & 13.78 & 58.78 & 11.76 & 97.86\\
CountSE & 2025 & Text & 0 & 8.51 & 54.93 & 7.84 & 82.99 \\
QICA & 2026 & Text & 0 & 12.98 & 56.35 & 12.41 & 97.28 \\
CounTR & 2022 & 3 visual exemplars & 3 & 13.13 & 49.83 & 11.95 & 91.23 \\
DAVE & 2024 & 3 visual exemplars & 3 & 8.91 & 28.08 & 8.66 & 32.36 \\
CountGD & 2024 & visual exemplars + text & 3 & 7.10 & 26.08 & 5.74 & 24.09 \\
CountingDINO & 2026 & 3 exemplars; training-free & 3 & 25.48 & 57.38 & 20.93 & 71.37\\
\hline
CounTR & 2022 & none  & 0 & 17.40 & 70.33 & 14.12 & 108.01\\
LOCA & 2023 & none & 0 & 17.43 & 54.96 & 16.22 & 103.96\\
DAVE & 2024 & none & 0 & 15.54 & 52.67 & 15.14 & 103.49 \\
UpCount (ours) & 2026 & none & 0 & \textbf{13.62} & \textbf{60.07} & \textbf{12.39} & \textbf{100.89}\\
\hline
\multicolumn{7}{l}{}
\end{tabular}
\end{table*}
\subsubsection{\textbf{FeatUp Refinement}}
We employ a FeatUp \cite{fu2024featup} mechanism to learn joint bilateral upsample (JBU) to recover fine spatial structures that may be lost during ViT patch tokenization. A lighweight convolutional stem first extracts a stride-four RGB guidance feature:
\begin{equation}
    \mathbf{F}_{\mathrm{RGB}}
    =
    \Phi_{\mathrm{RGB}}\left(\mathcal{X}_i\right)
    \in
    \mathbb{R}^{C_o \times 96 \times 96}.
\end{equation}
The projected final-block ViT feature is then upsampled from
$24 \times 24$ to $96 \times 96$ using learned joint bilateral kernels
conditioned on the RGB guidance:
\begin{equation}
    \mathbf{F}_{\mathrm{JBU}}
    =
    \Phi_{\mathrm{JBU}}
    \left(
        \widetilde{\mathbf{F}}^{(11)},
        \mathbf{F}_{\mathrm{RGB}}
    \right)
    \in
    \mathbb{R}^{C_o \times 96 \times 96}.
\end{equation}
Finally, the FeatUp representation is fused with the DPT feature and subsequently combined with the
RGB detail feature:
\begin{equation}
\begin{aligned}
    \mathbf{F}_{\mathrm{sem}}
    &=
    \mathbf{F}_{\mathrm{DPT}}
    +
    \Phi_{\mathrm{fuse}}
    \left(
        \mathbf{F}_{\mathrm{DPT}}
        \,\Vert\,
        \mathbf{F}_{\mathrm{JBU}}
    \right),\\
    \mathbf{F}_{\mathrm{ref}}
    &=
    \Phi_{\mathrm{detail}}
    \left(
        \mathbf{F}_{\mathrm{sem}}
        \,\Vert\,
        \mathbf{F}_{\mathrm{RGB}}
    \right),
\end{aligned}
\end{equation} where $\Vert$ denotes channel-wise concatenation and
$\mathbf{F}_{\mathrm{ref}}
\in\mathbb{R}^{C_o \times 96 \times 96}$ is the spatially refined
feature used by the subsequent counting modules.

\begin{table}[htbp]
\caption{Performance comparison on CARPK}
\label{tab2}
\begin{center}
\begin{tabular}{|c|c|c|c|}
\hline
\textbf{Method} & 
\textbf{Inference guidance}& \textbf{\textit{MAE}}& \textbf{\textit{RMSE}} \\
\hline
FamNet & visual exemplars & 18.19 & 33.66  \\
GMN & visual exemplars & 7.48 & 9.90 \\
BMNet+ & visual exemplars & 5.76 & 7.83 \\
CounTR & visual exemplars & 5.75 & 7.45 \\
SAFECount & visual exemplars & 5.33 & 7.04 \\
UpCount (ours)& none & \textbf{6.27} & \textbf{8.79} \\
\hline
\multicolumn{4}{l}{}
\end{tabular}
\end{center}
\end{table}

\subsubsection{\textbf{Proposal-Verification Counting}} The refined feature $\mathbf{F}_{\mathrm{ref}}$ is processed by a proposal-verification module. The module computes
non-local similarities between semantic ViT tokens and aggregates their
top-$K$ matches to identify repeated visual patterns. 
\begin{equation}
    \left(
        \mathbf{F}_{\mathrm{count}},
        \mathbf{S}_{\mathrm{rep}}
    \right)
    =
    \Phi_{\mathrm{rep}}
    \left(
        \widetilde{\mathbf{F}}^{(11)},
        \mathbf{F}_{\mathrm{ref}}
    \right),
\end{equation} where $\mathbf{F}_{\mathrm{count}}$ is the repetition-conditioned
counting feature and $\mathbf{S}_{\mathrm{rep}}$ represents the spatial repetition evidence.

A proposal head predicts a non-negative dense proposal map, while a
verification head evaluates each proposal using both the counting
feature and repetition signal:
\begin{equation}
\begin{aligned}
    \mathbf{P}_i
    &=
    \operatorname{Softplus}
    \left(
        \Phi_{\mathrm{prop}}
        \left(\mathbf{F}_{\mathrm{count}}\right)
    \right),\\
    \mathbf{V}_i
    &=
    \sigma
    \left(
        \Phi_{\mathrm{ver}}
        \left(
            \mathbf{F}_{\mathrm{count}}
            \,\Vert\,
            \mathbf{S}_{\mathrm{rep}}
        \right)
    \right),
\end{aligned}
\end{equation}
where $\mathbf{P}_i$ is the proposal density,
$\mathbf{V}_i$ is the verification map, and $\Vert$ denotes
channel-wise concatenation. Both maps are bilinearly upsampled to the
input resolution.

The verified proposal density is adjusted by a learned spatial
correction map $\mathbf{A}_i$ and a global count-calibration factor
$c_i$. The final density map and predicted count are given by
\begin{equation}
    \widehat{\mathbf{D}}_i
    =
    c_i
    \left(
        \mathbf{P}_i
        \odot
        \mathbf{V}_i
        \odot
        \mathbf{A}_i
    \right),
    \qquad
    \widehat{N}_i
    =
    \sum_{p}
    \widehat{\mathbf{D}}_i(p),
\end{equation}
where $\odot$ denotes element-wise multiplication.

\subsection{Two-Stage Training Scheme}
\subsubsection{Masked Autoencoder Pretraining} We first adapt the ViT-B/16 encoder to FSC-147 using masked image reconstruction. The encoder is initialized from the full ImageNet-MAE checkpoint and further pretrained using only the FSC-147 training split. The reconstruction loss is averaged across all patches:
\begin{equation}
\mathcal{L}_{\mathrm{MAE}}
=
\frac{1}{N}
\sum_{i=1}^{N}
\left\lVert \hat{\mathbf{x}}_i - \mathbf{x}_i \right\rVert_2^2
\end{equation} where $\mathbf{x}_i$ and $\hat{\mathbf{x}}_i$ denote the target and reconstructed RGB values of patch $i$.

\subsubsection{Supervised Counting Fine-Tuning} After pretraining, the MAE decoder is discarded, and the pretrained encoder is connected to the DPT pyramid, FeatUp refinement module, and proposal-verification counting head. The complete architecture, including all ViT blocks, is then optimized using density-map supervision. The supervised objective is the following.
\begin{equation}
    \mathcal{L}
    =
    \mathcal{L}_{\mathrm{density}}
    +
    \lambda_{c}\mathcal{L}_{\mathrm{count}}
    +
    \lambda_{\log}\mathcal{L}_{\mathrm{log\text{-}count}}
    +
    \lambda_{v}\mathcal{L}_{\mathrm{verify}}
\end{equation} where $\mathcal{L}_{\mathrm{density}}$ is a foreground-weighted density regression loss, $\mathcal{L}_{\mathrm{count}}$ enforces count-integral consistency, $\mathcal{L}_{\mathrm{log\text{-}count}}$ balances errors across different count ranges, and $\mathcal{L}_{\mathrm{verify}}$ supervises the verification response.

\begin{figure*}[htbp]
\centering
\includegraphics[width=\textwidth]{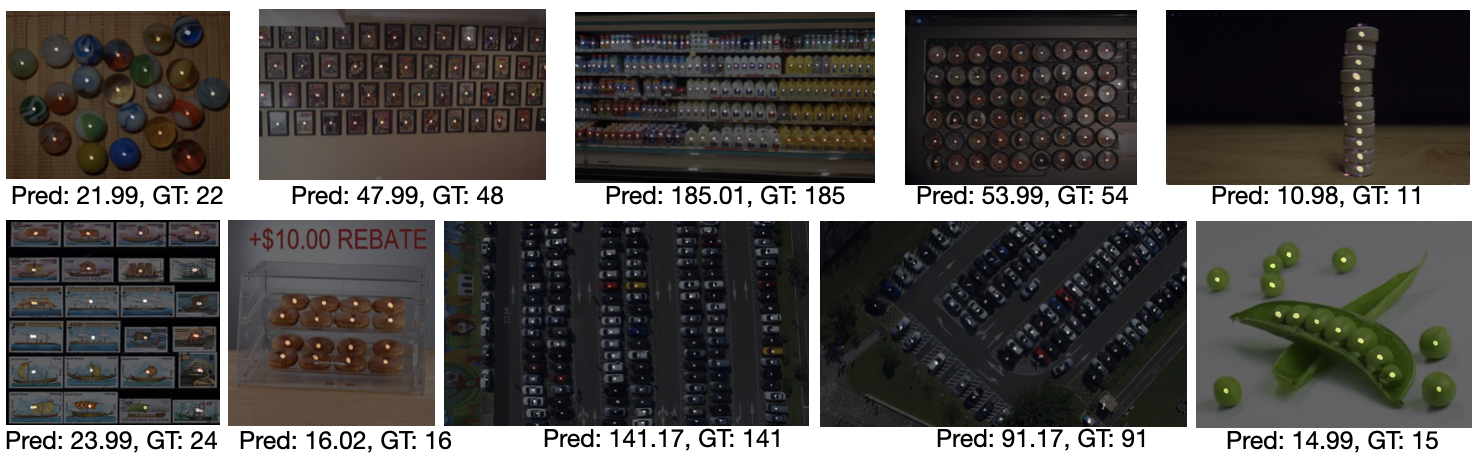}
\caption{Qualitative results of UpCount on FSC-147 and CARPK.}
\label{qual}
\end{figure*}

\section{Experiment \& Evaluation}
\subsection{Dataset \& Metrics}
\textbf{FSC-147.} 
The FSC-147 \cite{ranjan2021learning} dataset is widely used in class-agnostic counting and contains 6135 images spanning 147 object classes.
The number of objects per image ranges from 7 to 3731, with an average of 56.
FSC-147 provides a corresponding binary density map $y_i$ for each training image $X_i$ in $D_{train}$.

\textbf{CARPK.} We also utilize CARPK\cite{hsieh2017drone} dataset which consists of 1448 bird's-eye-view images of parking lots, with a total of 89,777 annotated cars.

\textbf{Metrics.} 
We use the standard Mean Absolute Error (MAE) and Root Mean Squared Error (RMSE) to evaluate the performance of the model.

\subsection{Implementation Details}
\textbf{Pretraining Details.} Pretraining is conducted for 500 epochs using AdamW, a batch size of 8, weight decay of 0.05, and a base learning rate of 1.5 $\times$ $10^{-4}$. The learning rate is linearly warmed up for 10 epochs and subsequently decayed using a cosine schedule.

\textbf{Supervised Fine-Tuning Details.} We perform separate 1000-epoch fine-tuning runs for FSC-147 and CARPK. Both datasets are initialized from the same FSC-147 pretrained MAE encoder; CARPK fine-tuning is not initialized from FSC-147 supervised fine-tuning.

FSC-147 fine-tuning uses a batch size of 26 and a base learning rate of 2 $\times$ $10^{-4}$, corresponding to an effective learning rate of approximately 2.03 $\times$ $10^{5}$. CARPK uses a batch size of 8 and an effective learning rate of 6.25 $\times$ $10^{-6}$. Both runs use AdamW, weight decay of 0.05, a 10-epoch linear warm-up, and cosine learning-rate decay, trained on a single NVIDIA RTX A6000 GPU.

\textbf{Inference Details.} Our model performs reference-less class-agnostic counting and does not require exemplar boxes, category labels, or textual queries at inference.

\subsection{Performance Comparison}
\textbf{Baselines.} We compare with representative counting methods covering different forms of inference guidance. For text-guided methods, we include T2ICount \cite{qian2025t2icount}, CountSE \cite{liu2025countse}, and QICA \cite{zhang2026boosting}. CounTR \cite{countr}, LOCA \cite{djukic2022low}, and DAVE \cite{pelhan2024dave} are included as representative few-shot and zero-shot counting methods. We also include CountGD \cite{amini2024countgd} and CountingDINO \cite{pacini2026countingdino} as multimodal, and training-free methods, respectively.

\textbf{Evaluation results.} Tab. \ref{tab1} compares UpCount with representative text-guided, exemplar-guided, training-free, and reference-free counting methods on FSC-147. Under this reference-free setting, it achieves an MAE of 13.62 on the validation set and 12.39 on the test set. UPCount also obtains the lowest test RMSE among the reference-free methods, reaching 100.89 compared with other baselines.

Tab. \ref{tab2} presents the results on the official CARPK test set following fine-tuning on its training set. UPCount achieves an MAE of 6.27 and an RMSE of 8.79 without requiring exemplar boxes during inference. These results indicate that the proposed representation transfers effectively from general class-agnostic counting to the more specialized vehicle-counting domain.

\textbf{Qualitative results.} We outline representative qualitative results in Fig. \ref{qual}, where UpCount produces coherent density maps on FSC-147 for cluttered scenes and structurally complex objects, and maintains stable car counts on CARPK.

\section{Conclusion}
In this work, we presented \textit{UpCount}, a class-agnostic object counter that improves counting on structurally complex objects by strengthening spatial feature representations. UpCount extracts multi-layer features from a ViT encoder, reassembles them into a multi-scale pyramid, and refines them using Dense Prediction Transformers and FeatUp before predicting a density map with a proposal--verification counting head. Experiments on FSC-147 demonstrate competitive performance in the reference-free setting, and results on CARPK indicate that the learned representation transfers effectively to vehicle counting.

\newpage
\section*{Acknowledgment}
This research is supported by Temasek Laboratories, Singapore University of Technology
and Design; the National Research Foundation, Singapore under its AI Singapore Programmes (AISG Award No.: AISG2-TC-2022-007); the Agency for Science, Technology
and Research (A*STAR) under its MTC Programmatic Funds (Grant No. M23L7b0021);
the National Research Foundation, Singapore and Infocomm Media Development Authority
under its Trust Tech Funding Initiative. Any opinions, findings and conclusions or recommendations expressed in this material are those of the author(s) and do not reflect the
views of National Research Foundation, Singapore and Infocomm Media Development Authority.




\bibliographystyle{IEEEtran}
\bibliography{IEEEexample}


\end{document}